\documentclass[11pt]{article}
\usepackage{coling2020}
\usepackage{times}
\usepackage{url}
\usepackage{latexsym}
\usepackage{graphicx}
\usepackage{booktabs}
\usepackage[compact]{titlesec}

\colingfinalcopy 

\title{What the [MASK]?\\Making Sense of Language-Specific BERT Models}

\author{Debora Nozza, Federico Bianchi \and Dirk Hovy \\
        Bocconi University \\ Via Sarfatti 25, 20136 Milan \\
        \texttt{\{debora.nozza, f.bianchi, dirk.hovy\}@unibocconi.it}
        }

\date{}

\begin{document}
\maketitle
\begin{abstract}

Recently, Natural Language Processing (NLP) has witnessed an impressive progress in many areas, due to the advent of novel, pretrained contextual representation models. 
In particular, \newcite{devlin2018bert} proposed a model, called BERT (\textbf{B}idirectional \textbf{E}ncoder \textbf{R}epresentations from \textbf{T}ransformers), which enables researchers to obtain state-of-the art performance on numerous NLP tasks by fine-tuning the representations on their data set and task, without the need for developing and training highly-specific architectures.
The authors also released multilingual BERT (mBERT), a model trained on a corpus of 104 languages, which can serve as a universal language model. This model obtained impressive results on a zero-shot cross-lingual natural inference task. 
Driven by the potential of BERT models, the NLP community has started to investigate and generate an abundant number of BERT models that are trained on a particular language, and tested on a specific data domain and task. This allows us to evaluate the true potential of mBERT as a universal language model, by comparing it to the performance of these more specific models. 
This paper presents the current state of the art in language-specific BERT models, providing an overall picture with respect to different dimensions (i.e. architectures, data domains, and tasks). Our aim is to provide an immediate and straightforward overview of the commonalities and differences between Language-Specific (language-specific) BERT models and mBERT. 
We also provide an interactive and constantly updated website that can be used to explore the information we have collected, at \url{https://bertlang.unibocconi.it}.
\end{abstract}

\section{Introduction}
\label{intro}

%
%

In all natural languages, word meaning varies with and is determined by context, and one of the main challenges of Natural Language Processing (NLP) has been (and remains) to model this property of meaning. Embedding-based language models \cite{mikolov2013distributed} have been shown to capture word meaning more efficiently than previous methods, allowing for both qualitative analysis of similarities and improved performance when used as input to predictive models. However, while embeddings represent word \textit{types} based on their general contextual co-occurrences, they do not learn context-specific representations for each word \textit{token}.

Recently, NLP has witnessed the advent of a groundbreaking new language model developed by Google researchers, called Bidirectional Encoder Representations from Transformers (BERT)~\cite{devlin2018bert}. It learns contextual representations for word tokens, thereby getting at their contextual variation in meaning. Contextualized BERT embeddings have since also dominated the leaderboards in a wide variety of NLP tasks.

The power of BERT representations lies in the fact that it is essentially a pretrained model that can be fine-tuned over specific downstream tasks, which enables it to achieve state-of-the-art results. The fundamental underlying component of this architecture is the Transformer model~\cite{vaswani2017attention}, an attention-based mechanism that has been shown to be effective in many different tasks. Both the Transformer and BERT have gathered much attention, and there is now a wealth of research articles and blog posts describing the inner workings of these models~\cite[among others]{rogers2020primer}.

Given the overwhelming success of BERT, a multilingual BERT model (mBERT)\footnote{\url{https://github.com/google-research/bert/blob/master/multilingual.md}} has been proposed, supporting over 100 languages, including Arabic, Dutch, French, German, Italian, or Portuguese. The model is trained on different domains, like social media posts or newspaper articles. mBERT has shown great capabilities in zero-shot cross-lingual tasks~\cite{pires2019multilingual}.

Due to the remarkable results of these models, an abundant number of BERT model extension has recently been introduced by researchers and industry practitioners from several countries: Currently, there are around 5k repositories mentioning ``bert'' on GitHub.com, and we can expect further demand for BERT extensions. 
These models are trained on a particular language and tested on a specific data domain and task, with the promise of maximizing performance across more tasks in that language, saving other users further fine-tuning.

However, it has so far not been clearly demonstrated whether the advantage of training a language-specific model is worth the expense in terms of computational resources\footnote{These models require a large amount of computational resources unaffordable for many users, and comes with severe ecological costs: training BERT on a GPU is roughly equivalent to a trans-American flight in terms of CO2 emissions~\cite{strubell2019energy}).}, rather than using the \textit{unspecific} multilingual model. 

Moreover, the NLP community is now facing a problem organizing the plethora of models that are being released. These models are not only trained on different data sets, but also use different configurations and architectural variants. To give a concrete example, the original BERT model was trained using the WordPiece tokenizer~\cite{wu2016google}, however, a recent language-specific model (CamemBERT~\cite{martin2019camembert}) used the SentencePiece tokenizer (available as OSS software)~\cite{kudo2018sentencepiece}.

Identifying which model is the best for a specific task, and whether the mBERT model is better than language-specific models is a key step future progress in NLP, and will impact the use of computational resources. 
Surveying both GitHub and the literature, we identified 30 different pretrained language-specific BERT models, covering 18 Languages and tested on 29 tasks, resulting in 177 different performance results~\cite{le2019flaubert,antoun2020arabert,martin2019camembert,alabi2019massive,kuratov2019rubert,slavicbert,virtanen2019finbert,polignano2019alberto,de2019bertje,cui2019chbert}. We outline some of the parameters here, and introduce the associated website for up-to-date searches.
We hope to give  NLP researchers and practitioners a clear overview of the tradeoffs before approaching any NLP task with such a model.\\

The contributions of this paper are the following:
\begin{enumerate}
    \item we present an overall picture of language-specific BERT models from an architectural, task- and domain-related point of view;
    \item we summarize the performance of language-specific BERT models and compare with the performance of the multilingual BERT model (if available);
    \item we introduce a website to interactively explore state-of-the-art models. We hope this can serve as a shared repository for researchers to decide which model best suits their needs.
\end{enumerate}

\section{Bidirectional Encoder Representations from Transformers}
We assume that most readers who are interested in the topic have a basic understanding of BERT and its components. However, for completeness' sake, we include a brief and high-level overview of the most important aspects here.

\subsection{BERT}

BERT uses the Transformer \cite{vaswani2017attention} architecture to learn word embeddings. The Transformer is a recent architectural advancement that can be included in deep networks for sequence modeling. Instead of modeling sequences as RNNs or LSTMs, the Transformer learns global dependencies between input and output, using only attention mechanisms.

Transformers greatly shifted the focus of the research community towards attention-based architectures. The encoder-decoder structure based on transformers is also incorporated into BERT.

\begin{figure}[t]
    \centering
    \includegraphics[width=\linewidth]{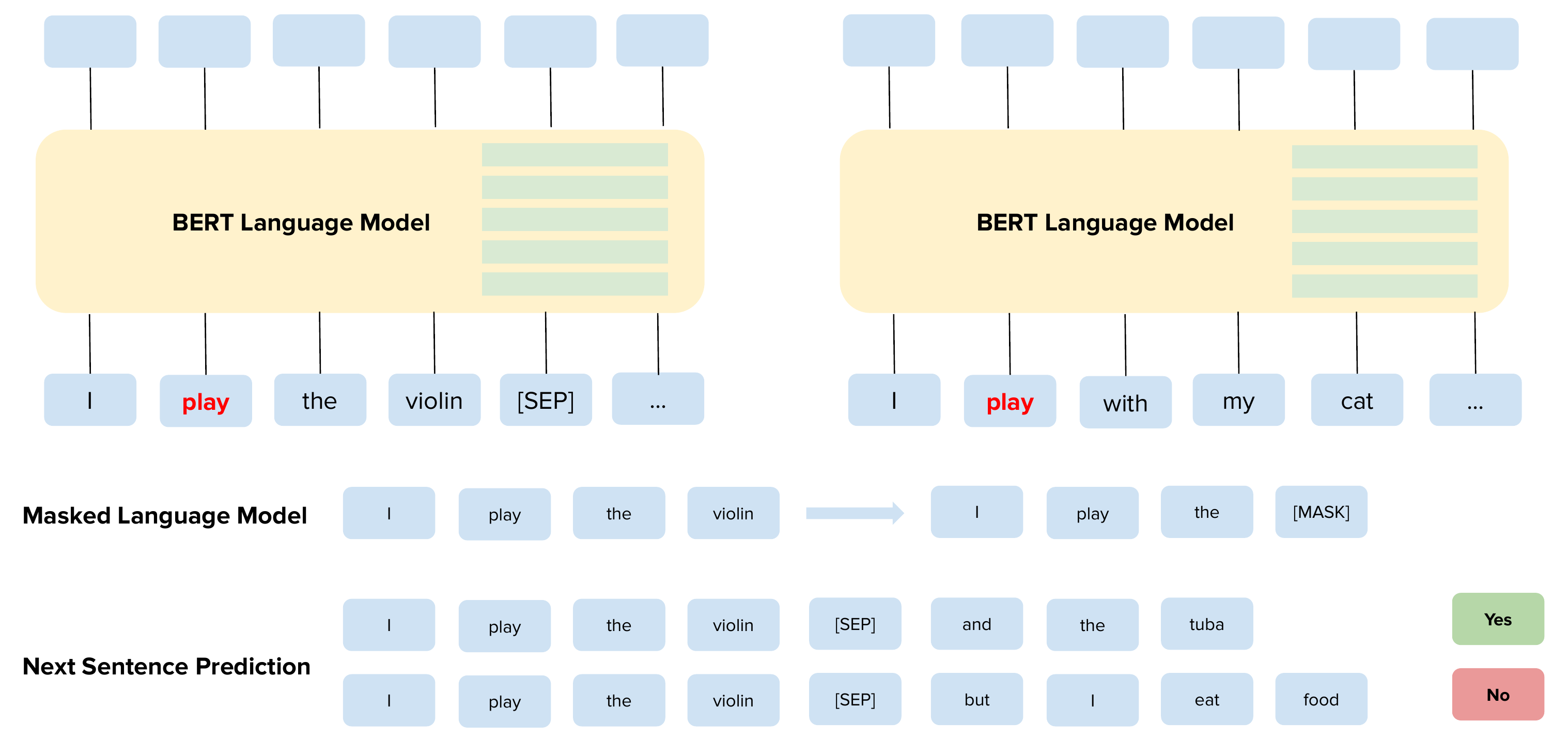}
    \caption{A schematic representation of BERT, masked language model and next sentence prediction. Different words have different meanings  and BERT looks at the word context to generate  contextual representations.}
    \label{fig:bertino:picture}
\end{figure}

\newcite{devlin2018bert} introduced BERT in 2018 as a context-sensitive alternative to previous word embeddings (which assume a word always has the same representation, independent of its context). The model essentially stacks several encoder-decoder structures based on transformers together. It uses masks to blank out individual words, forcing the model to ``fill in the blanks'', thereby increasing its context-sensitivity. 
Two key elements in the BERT pretraining process are the masked language model and the next sentence prediction. In the former process, a random subsample (in the BERT paper, 15\%) of the words in a text are replaced by a \texttt{[MASK]} token, and the task is to predict the correct token. 
The latter process instead is the task of predicting how likely one sentence is to follow another one in text. See Figure~\ref{fig:bertino:picture} for a schematic view on BERT.
Other than traditional word embeddings, BERT representations are not a fixed lookup table, but require the full context to produce a word representation. The vocabulary is defined in advance and it is based on WordPiece~\cite{wu2016google}, a tokenization algorithm that generates sub-word tokens.

Due to its size in terms of parameters, the model usually comes in a pretrained format, which can be fine-tuned on the task or data set. Simple classification layers can be stacked on top of the pretrained BERT to provide predictions for several tasks such as sentiment analysis or text classification.

{}

\subsection{Multilingual BERT, ALBERT and RoBERTA}
Subsequently, BERT was extended to include several languages. Multilingual (mBERT) was part of the original paper \cite{devlin2018bert}, and is pretrained over several languages using Wikipedia data.  
This allows for zero-shot learning across languages, i.e., training on data from one language and applying the model to data in another language.

Along the same lines, \newcite{lan2019albert} introduced A Lite BERT (ALBERT), to reduce the computational needs of BERT. ALBERT includes two parameters reduction techniques, that reduce the number of trainable parameters without significantly compromising performance. Moreover, the authors introduce another self-supervised loss, related to sentence order prediction that is meant to address the limits of next sentence prediction used in the BERT model. Another recent paper~\cite{liu2019roberta} has shown that BERT is sensitive to different training strategies; the authors introduce RoBERTA~\cite{liu2019roberta} as a well-optimized version of BERT.

\section{Making-Sense of Language-Specific BERT Models}

While multi- and cross-lingual BERT representations allow for zero-shot learning and capture universal semantic structures, they do gloss over language-specific differences. Consequently, a number of language-specific BERT models have been developed to fill that need. These models almost always showed better performance on the language they were trained for than the universal model.

In order to navigate this wealth of constantly changing information, a simple overview paper is no longer sufficient. While we aim to give a general overview here, we refer the interested reader to the constantly updated online resource, BertLang.

\subsection{BertLang}

We introduce BertLang (\url{https://bertlang.unibocconi.it}), a website where we have gathered different language-specific models that have been introduced on a variety of tasks and data sets. 
Most of the models are available as GitHub links, and some of them are described in research papers, but very few have been published in peer-reviewed conferences\footnote{We do not include resources that feature only a model without reporting any performance results.}. 
In addition to providing a searchable interface, BertLang also provides the possibility to add new information
While we hope to independently verify the reported results in the future, for now, we only list the various models and conditions.

We open-source both data and code to build the website\footnote{\url{https://github.com/MilaNLProc/bertlang}}, this will make it possible for other researchers to contribute to the collection of language-specific BERT models.

\begin{figure}[t]
    \centering
    \includegraphics[width=\linewidth]{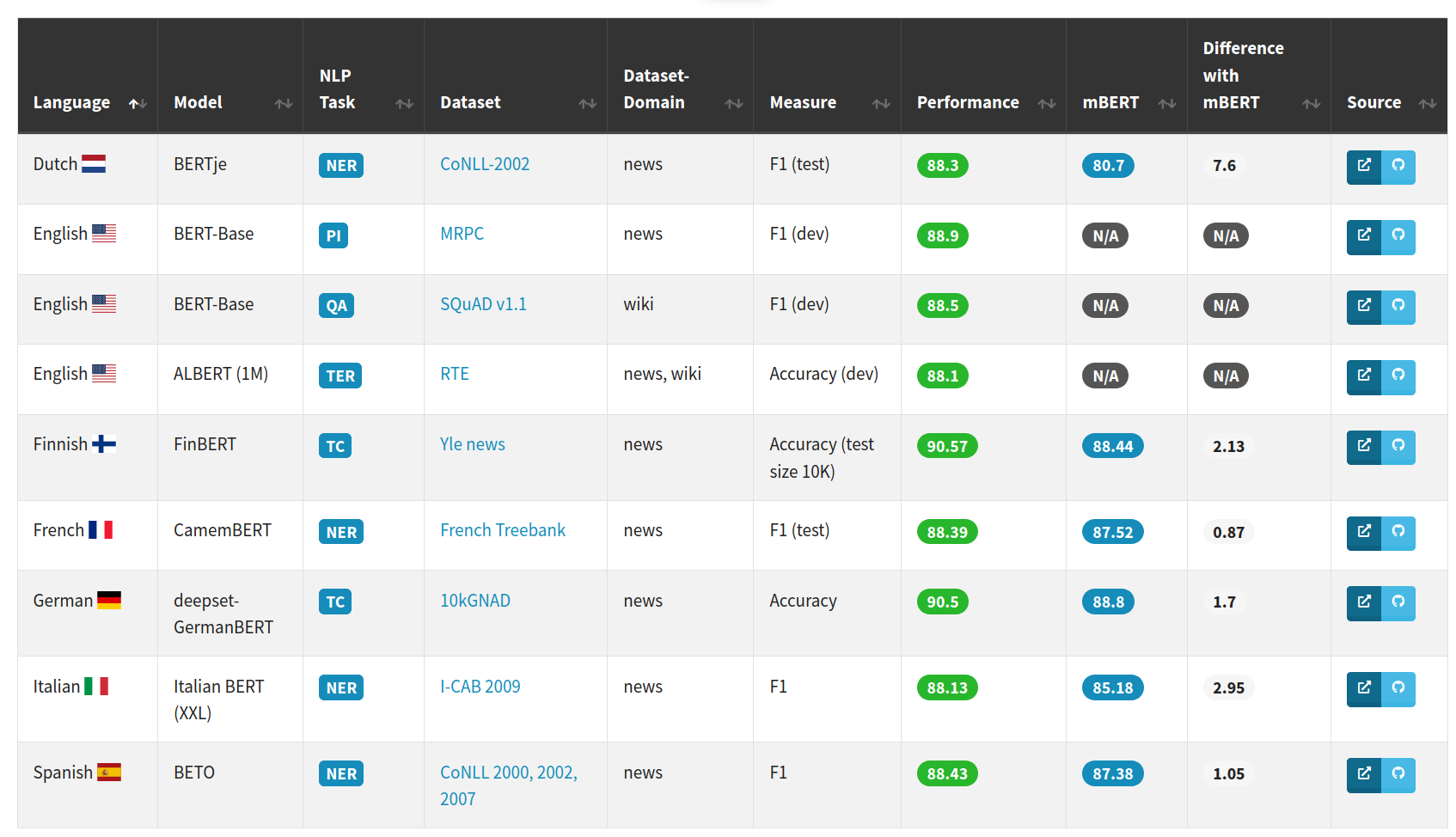}
    \caption{The BertLang website front-end interface.}
    \label{fig:bertlang:picture}
\end{figure}{}

Figure~\ref{fig:bertlang:picture} shows the frontend page of our website, showing a table that contains languages, tasks, and performances of different models. We also provide links to the references and code from which we retrieved that information. Beyond this information, we report the performance evaluation metric, the average performance obtained by the language-specific, and -- where available -- the corresponding performance of mBERT model and their difference.


\begin{table}[]
    \centering
    \begin{tabular}{l|c|c|c|c} \toprule
         \textbf{Task} & \textbf{Metric} &\textbf{Avg. lang-specific BERT} & \textbf{Avg. mBERT} & \textbf{Diff.}  \\ \midrule
         Named Entity Recognition & F1 & 85.26 & 80.87 & 4.39 \\ 
Natural Language Inference & Accuracy & 78.35 & 74.60 & 3.75 \\ 
Paraphrase Identification & Accuracy & 88.44 & 87.74 & 0.70 \\ 
Part of Speech Tagging & Accuracy & 97.06 & 95.87 & 1.19 \\
Part of Speech Tagging & UPOS & 98.28 & 97.33 & 0.95 \\ 
Sentiment Analysis & Accuracy & 90.17 & 83.80 & 6.37 \\ 
Text Classification  & Accuracy & 88.96 & 85.22 & 3.75 \\ \bottomrule
    \end{tabular}
    \caption{Summary of average performance of different language-specific BERT models on various tasks.}
    \label{tab:summary:nlp:tasks}
\end{table}{}

\subsection{Language-Specific BERT Models}
The models we index vary along with a number of dimensions, which we discuss below. The main distinction, however, is the specific language the model was trained on. The availability of data sets in that language determines the tasks and domains this model was applied to.

Table \ref{tab:summary:nlp:tasks} shows a summary of the results for the most frequent NLP tasks investigated across several languages. 
The results clearly show that on average, language-specific BERT models obtain higher results with respect to mBERT in all the considered tasks. However, while this holds for averages, with the proliferation of languages, tasks, and data sets, there is a huge variation in the individual performances. In the following, we analyze the possible views of the collected results in more detail.

\paragraph{Languages Covered}
The language-specific BERT models proposed range from languages that have a high number of resources available on the web for training (e.g., French, Italian) to low-resource languages, such as Yorùbá and Mongolian. At the current date, we are covering 18 languages.

Interestingly, from the results it is possible to grasp that low-resources languages (e.g., Yorùbá and Arabic) are actually the ones with the highest improvement with respect to mBERT. Since mBERT is trained on Wikipedia, this finding can probably be explained by the fact that developers of language-specific BERT models are more likely to be experts on other resources for that language, or to collect more data. This makes a greater difference for low-resource languages.

\paragraph{Architectures}
The most popular architecture is the standard BERT one, but lately, the introduction (and the good performances) of both ALBERT and RoBERTA has made researchers consider those two latter models as well to pretrain language models.

RoBERTA has been used as the base model for the French CamemBERT\cite{martin2019camembert}, as well as the Italian Gilberto\footnote{\url{https://github.com/idb-ita/GilBERTo}} and Umberto\footnote{\url{https://github.com/musixmatchresearch/umberto}}. 

mBERT was used to initialize and fine-tune models for languages such as Russian~\cite{kuratov2019rubert}, Slavic languages~\cite{slavicbert}\footnote{Here ``Slavic'' includes Russian, Bulgarian, Czech and Polish.} and Yorùbá~\cite{alabi2019massive}. The latter is a noteworthy example of how the scarcity of available data in low resource languages can be overcome. Fine-tuning mBERT instead of pretraining from scratch allowed the authors to produce a model without access to large amounts of data.

\paragraph{NLP tasks}
We currently index results for 29 NLP tasks. Table \ref{tab:summary:nlp:tasks} reports the results for the most popular tasks in the collected data, with Named Entity Recognition (NER) the most frequent task (22~entries). 
Looking at the source of the test data (see the released website for the complete information), we observe that there are some multilingual benchmark data sets that are used for the same NLP task in different languages. Some of them have been released by research group publishing in well-known NLP conferences \cite{pawsx,partut,xnli,volker2019hdt}, while others have been released in conjunction with shared tasks such as SemEval or CoNLL \cite{conll18,navigli13semeval,bosco2016evalita,benikova2014germeval}. The latter group shows the effect shared task have on providing the NLP community with benchmark references.

Remarkably, the noun sense disambiguation task is the only task where language-specific BERT performances are lower than the mBERT ones. As stated by the authors \cite{le2019flaubert}, this could be due to the fact that the training corpora have been machine-translated from English to French, making mBERT probably better suited for the task than a model trained on native French.

Sentiment analysis is the task where language-specific BERT models obtain the highest improvements with respect to mBERT. Following the previous intuition, for Arabic \cite{antoun2020arabert} this can be explained by considering the peculiar language of the test data set, which demonstrates the ability of the language-specific AraBERT model to handle dialects --- even if they were not explicitly included in the training set.

Beyond the well-known NLP tasks, it is interesting to note that language-specific tasks have been investigated as well, e.g., the Die/Dat (gendered determiners) disambiguation task in Dutch \cite{delobelle2020robbert}, obtaining impressive improvements with respect to state-of-the-art \cite{allein2020binary} ($\sim$~23\% points accuracy improvement).

\paragraph{Domains}
There is a huge variety of domains considered in language-specific BERT models. We need to make a distinction, though, between data sets used to pretrain the models and data sets used to evaluate the models. 

Data used for training mainly varies across three source corpora: (i) Wikipedia, (ii) OPUS Corpora~\cite{tiedemann2012parallel} and (iii) OSCAR~\cite{ortizsuarez:hal-02148693}.
Wikipedia is currently comprising more than 40 million articles created and maintained as an open collaboration project in 301 different languages, making it the largest and most popular multilingual online encyclopedia. mBERT, for example, was trained over 100 different language-specific Wikipedia versions.
OPUS is a freely available collection of parallel corpora, covering over 90 languages. The largest domains covered by OPUS are legislative and administrative texts, translated movie subtitles and localization data from open-source software projects~\cite{tiedemann2012parallel}.
OSCAR (Open Super-large Crawled Almanach coRpus)~\cite{ortizsuarez:hal-02148693} is a huge multilingual corpus obtained by filtering the Common Crawl corpus, which is a parallel multilingual corpus comprised of
crawled documents from the internet.

 Several models concatenate more sources to have enough data to pretrain BERT, for example BERTje (Dutch BERT), which concatenates news, book data, and Wikipedia data and other text. Languages with more limited availability of data, such as Yorùbá, have brought researchers to fine-tune mBERT instead of pretraining from scratch. A notable case is the Italian BERT model ALBERTO \cite{polignano2019alberto}, which is the only one that has been trained only on social media data (specifically, on 2~million Twitter posts in Italian language).

On the other hand, different domain data sets have been used to evaluate the models; these range from review data for sentiment analysis tasks to transcripts and news for more traditional tasks, such as part of speech tagging.
News data are the most common domain, presumably because they are easier to retrieve, and because their more formal register makes them more suited for tasks such as part of speech tagging, dependency parsing, and named entity recognition. 
Similarly, social media posts from Twitter are mostly used in tasks like sentiment analysis and identification of offensive language.

\section{Conclusions}

BERT~\cite{devlin2018bert} has greatly improved results in many different NLP tasks and has become a mainstay of the community. Following this development, a multilingual BERT and several language-specific versions have been developed and contributed even more to the success of NLP applications.

In this paper, we have analyzed the current state-of-the-art, showing languages are covered, which tasks tackled, and which domains considered in pretrained language-specific BERT models. Moreover, we have underlined the huge variability models and the difficulty for researchers to find the best model for a specific task, language, and domain. To this end, we have introduced BertLang, a website that allows researchers to search and explore the current state-of-the-art with respect to language-specific BERT models.

In the future, we plan to provide independent verification of reported results and direct comparisons of language-specific BERT models on specific domains and tasks. We plan to use the same data to fine-tune the models providing comparable performance values for the models. We believe these comparisons will be beneficial to the community of both researchers and beginning practitioners in NLP.




\bibliographystyle{coling}
\bibliography{coling2020}

\end{document}